# A Belief-Function Based Decision Support System

Hong Xu*†            Yen-Teh Hsia*            Philippe Smets*

*IRIDIA-Université libre de Bruxelles
50, Ave. F. D. Roosevelt, CP 194/6, B-1050, Brussels, Belgium

†School of Business, University of Kansas
Lawrence, Kansas 66045, USA

## Abstract

In this paper, we present a decision support system based on belief functions and the pignistic transformation. The system is an integration of an evidential system for belief function propagation and a valuation-based system for Bayesian decision analysis. The two subsystems are connected through the pignistic transformation. The system takes as inputs the user's "gut feelings" about a situation and suggests what, if any, are to be tested and in what order, and it does so with a user friendly interface.

## 1. INTRODUCTION

Decision making under uncertainty is a common problem in the real world. Decision analysis provides a method for decision making. The main objective of this method is to help the decision maker to select an appropriate decision alternative in the face of uncertain environment. Traditional Bayesian decision analysis is based on Bayesian probability theory and utility theory, where uncertainty in the states of nature are represented by probabilities. Some popular methods for representing and solving Bayesian decision problems are decision trees and influence diagrams.

Recently, a unified framework for uncertainty representation and reasoning, called valuation-based system (VBS), has been proposed (Shenoy 1989, 1991b). It can represent knowledge in different domains including probability theory, Dempster-Shafer theory and possibility theory. More recent studies show that the framework of VBS is also sufficient for representing and solving Bayesian decision problems (Shenoy 1991a, 1992). The graphical representation is called a valuation network, and the method for solving problems is called the fusion algorithm. Shenoy (1992) has shown that the representation and solution method of VBS is more efficient than that of decision trees and that of influence diagrams.

Dempster-Shafer theory (Shafer 1976, Smets 1988) aims to model a decision maker's subjective valuation of evidence. It allows one to express partial beliefs when sufficient information is not available. Some methods for using belief functions for decision analysis have been studied, such as Jaffray (1989), Yager (1989), Smets (1990) and Strat (1990). As the particular power of VBS is its applicability to many different uncertainty calculi, we have proposed two methods for decision analysis using belief functions in the framework of VBS: One is to generalize the framework of VBS for Bayesian decision analysis to accept belief function representation and computation (Xu 1992a), the other is based on the transferable belief model (Smets 1990). In this paper, we will present a decision support system based on the transferable belief model (TBM). The system is an integration of two subsystems: an evidential system for belief function propagation and a valuation-based system for Bayesian decision analysis, which are connected through the pignistic transformation as described in the context of the TBM. We will give an example concerning a nuclear waste disposal problem to demonstrate the use of the system.

The remainder of the paper is as follows: In section 2, we briefly review the TBM. In section 3, we describe the structure and the algorithm of the Decision Support System based on the TBM. In section 4, we give an example to show how to use the system for decision making in the real world. Finally, in section 5, we present our conclusions.

## 2. THEORETICAL BACKGROUND - TRANSFERABLE BELIEF MODEL

The transferable belief model (TBM) is a model developed to represent someone's degree of beliefs. This model is based on the use of belief functions and is closely related to the model that Shafer (1976) has described. Smets and Kennes (1990) described the TBM and compared it with the classical Bayesian model. The TBM is based on:
- a two-level structure: the **credal** level where beliefs are entertained and the **pignistic** level where beliefs are used to make decisions;
- beliefs at the credal level are quantified by **belief** functions;
- beliefs at the pignistic level are quantified by **probability** functions.
- when a decision must be made, beliefs at the credal level are transformed into beliefs at the pignistic level, i.e. there exists a transformation from belief functions to probability functions.

Decision making requires that we derive a probability function that can be used to compute expected utilities of each potential decision. It means that uncertainty at the pignistic level must be quantified by a probability function. But it does not mean that beliefs at the credal level must also be quantified by a probability function. What is required is that there exists some transformation between the representation at the credal level and the probability function that must exist at the pignistic level. This problem in the context of the TBM can be solved by imposing some rationality requirement that leads to the concept of the pignistic transformation (Smets 1990). Hence, when decision must be taken, the TBM is endowed with the needed procedure to transform someone's beliefs entertained at the credal level into a so-called pignistic probability that can be used at the pignistic level. The justification of the pignisitic transformation is based on rationality, normative requirements (Smets and Kennes, 1990). If m is the basic belief masses on a space $\Omega$, then for every element $\omega$ of $\Omega$, the pignistic probability:

$$BetP(\omega) = \sum_{A: \omega \in A \subseteq \Omega} \frac{m(A)}{|A|}$$ where $|A|$ is the number of elements of $\Omega$ in A.

## 3. A BELIEF FUNCTION BASED DECISION SUPPORT SYSTEM

Based on TBM described above, we developed a belief function based decision support system. The system architecture is illustrated in Figure 1. The core of the system is a two-level structure: one called TRESBEL (a Tool for REaSoning with BELief functions), the other called VBSD (a Valuation-Based System for Decision analysis). They correspond to the two-level structure of the TBM: the credal level and the pignistic level. Users interact with the system through the user interface.

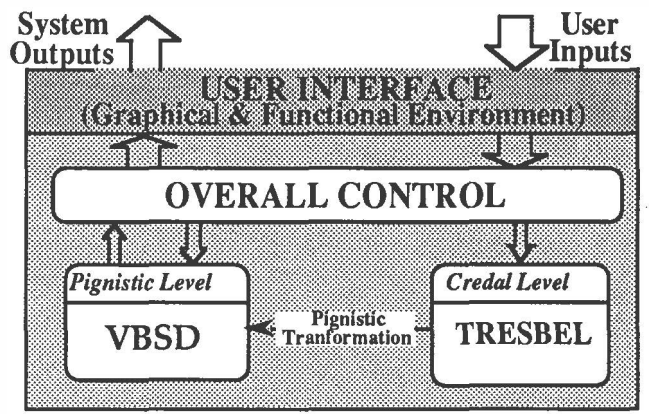

Figure 1: System Architecture

The system takes as inputs the user's "gut feelings" (presented as belief functions) about situations through a user-friendly interface, propagates the belief functions in TRESBEL, compute the pignistic probabilities needed for decision making, transfers these pignistic probabilities into the VBSD for decision analysis, and suggests the optimal decision through the User Interface. The whole procedure is controlled by the module called overall control.

The use of our system is presented within a framework where tests may be done, symptoms are observed, diagnosis must be established, and treatments must be selected.

Through the user interface, users have to provide the following input information to construct model:
- A list of tests (default results: +/- or positive/negative) among which the system must select the most appropriate one to be performed next given those already performed.
- A list of observed symptoms (default values: yes/no);

- One diagnosis node (no default; user has to explicitly provide the possible diagnoses);
- A list of potential treatments (=actions);
- A utility matrix (utility of the treatment given the diagnosis);
- For each test, specify the corresponding cost;
- Specify "gut feelings" as a bunch of belief functions relating the tests and the symptoms to the diagnosis, as well as the a priori on the diagnosis (it will be represented by the vacuous belief function when there is no a priori).

As system outputs, users can get the following information:
- A tree of suggested tests (depth of tree can be defined by users), including "no test" as an alternative;
  - The tree can be visually adjusted;
  - For each node, the user can require: (1) why the test is suggested; (2) what is the ranking of the treatments at the moment.

The algorithm for constructing a tree of suggested tests is as follows:

**Build-Tree**(CurrentTree)
  Step 1: Given the current specification (a bunch of belief functions), first compute the belief (BF) on (i) the diagnosis variable and (ii) each test variable, then transform each BF into its corresponding pignistic probability (BetP).
  Step 2.1: Given the BetP on the diagnosis variable, determine the optimal treatment, i.e. the one with the largest expected utility. We call this treatment and its associated utility MaxU(0) (the index 0 corresponds to 'no test')
  Step 2.2: For each test i that has no yet been performed, set the test result as positive (by setting a belief function with a basic belief mass = 1 on "+") and negative, respectively, and get the corresponding best treatment. Let MaxU(i+) and MaxU(i-) be the utility expected from the best treatment when test i outcome is positive and negative, respectively. Let MaxU(i) be the utility of test i, and BetP(i+) and BetP(i-) be the pignistic probabilities of the test i outcomes. Then,
  MaxU(i)=BetP(i+)·MaxU(i+)+BetP(i-)·MaxU(i-)
          - Cost(i).
  The largest MaxU(i), the best the test.
  Step 3. Among all the tests considered in step 2, select the one that has the largest expected utility MaxU(i), make it the root of the current tree, and record the corresponding treatment (also record the reason why this test is selected, as well as the ranking and the utilities of all the alternative treatments at the moment). Suppose the selected test is test j. IF test j = "no test" or the depth of the current tree is not less than the specified depth, THEN go to step 5; ELSE
  Step 4. Set test j as "performed test". In order to find the next optimal test given the tests already selected, set result of test j as positive; call **Build-Tree** (right child); set result of test j as negative, call **Build-Tree** (left child).
  Step 5. Stop computation and display the tree.

At the credal level, we use a tool called TRESBEL for computing BF on (i) and (ii) of Step 1. TRESBEL is a software developed at IRIDIA (Xu 1992b) for propagating belief functions in belief networks (Zarley et al. 1988, Hsia and Shenoy 1989, Xu 1991b). It is based on local computation techniques (Shafer and Shenoy 1988). Especially, it carefully tackles the issue of efficiency (Xu 1991a). The reason we use TRESBEL is its distinct feature of performing fast computation and incremental changes.

At the pignistic level, we use a tool called VBSD, an extension of Pulcinella (Saffiotti & Umkehrer 1991) for computing maximum expected utility. VBSD is another implementation being developed at IRIDIA for Bayesian Decision Analysis in VBS. It is based on the fusion algorithm for solving Bayesian decision problem proposed by Shenoy (1991a, 1992). Shenoy (1992) has shown that the VBS representation and solution method is more efficient than some traditional methods such as decision trees and influence diagrams.

## 4. AN EXAMPLE

In this section, we give an example of nuclear waste disposal to illustrate the use of our system.

In Figure 2, there is a river in a delta with two arms. Some radioactive product is leaking from some waste repositories located around the river. The leaking repository might be one of three known underground dumps (a, b, or c), or of four known truck dumps (d, e, f, or g). The leakage might also occur at a depository for which location we do not know and denoted as ω? (we don't know if it exists, or if it exists we don't know its position). We must find the location of the leakage and then clean it. To this end, we could make tests at the seven known locations (numbered 15 to 21 in Figure 2) and at some points along the river (denoted by the crosses in Figure 2, numbered 1 to 14). There is also a reservoir that may be contaminated and tested. The costs of the tests are known.

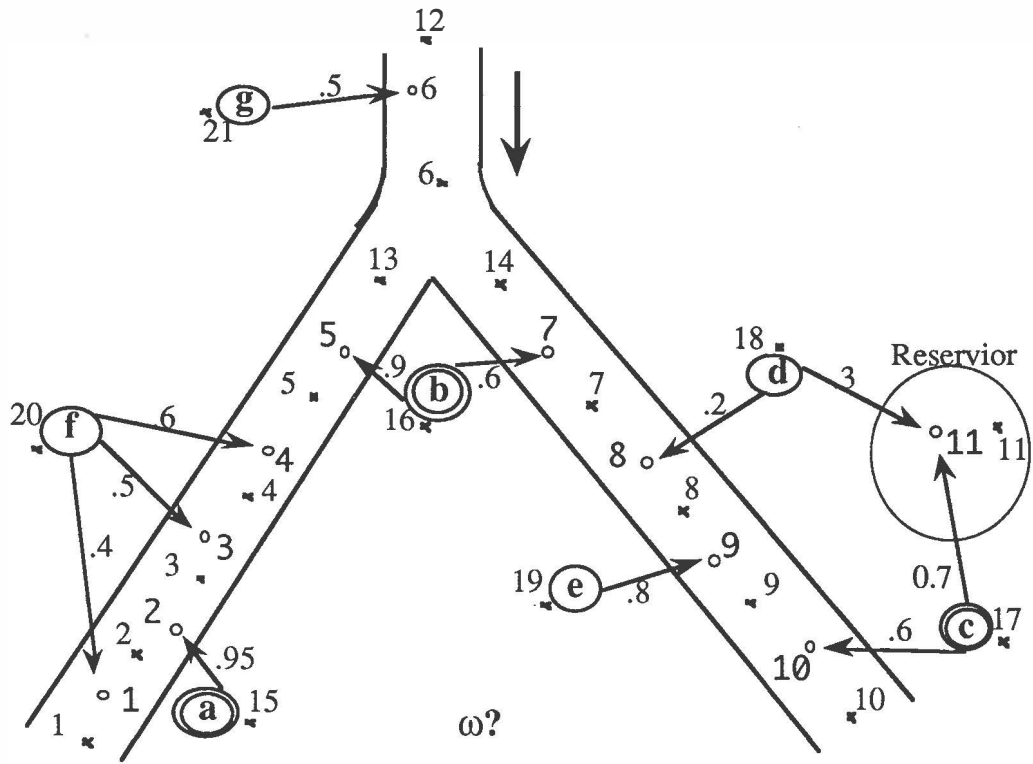

Figure 2: Example of nuclear waste disposal

The experts can express their beliefs that some location along the rivers is contaminated given the location of the leakage. This information is presented in Figure 2 by arrows. The weighted arrows in Figure 2 means: If the radioactive product is leaking from the location at the bottom of the arrow, then the belief that the contamination could reach the point indicated by the end of the arrow is quantified by a simple support function. The weight on the arrow is the basic belief mass given to the fact that the contamination has reached the considered potential location of contamination.

We also have the following constraints:
(1) IF point j is contaminated, THEN test j down the point has positive result, with m(test j positive) = .99;
(2) IF test i is positive, THEN test j next to it (down the river) has positive result, with m(test j positive) = .9;
(3) Test result at a depository location is positive given that the radioactive product is leaking from there, with m(test positive) = .99.

The costs of the tests are (k$):
cost for each test on the river (test-1 to test-14 except test-11) is 1;
cost for test-11 (in the reservoir) is 2;
cost for the tests at location a to g is 5, 7, 3, 2, 3, 3, and 4, respectively.

The payoffs for the treatments (digging to the leak) given the leaking location are as below:

|  |  | origin of radiation: | | | | | | | |
|---|---|---|---|---|---|---|---|---|---|
|  |  | pit | pit | pit | truck | truck | truck | truck | pit |
| cost (dig) | act (dig) | a | b | c | d | e | f | g | ω? |
| 50 | a | -50 | -250 | -450 | -100 | -100 | -100 | -100 | -150 |
| 60 | b | -360 | -60 | -460 | -110 | -110 | -110 | -110 | -160 |
| 60 | c | -360 | -260 | -60 | -110 | -110 | -110 | -110 | -160 |
| 10 | d | -310 | -210 | -410 | -10 | -60 | -60 | -60 | -110 |
| 10 | e | -310 | -210 | -410 | -60 | -10 | -60 | -60 | -110 |
| 15 | f | -315 | -215 | -415 | -65 | -65 | -15 | -65 | -115 |
| 10 | g | -310 | -210 | -410 | -60 | -60 | -60 | -10 | -110 |
| cost delay | no-clean | 300 | 200 | 400 | 50 | 50 | 50 | 50 | 100 |

The payoff of an action given the leaking position is the sum of the digging cost at that position and the cost of delay that would result if the leaking position is not the position being explored. Indeed, if one explores the wrong position, the cleaning of the leakage will be delayed and the cost of delay quantifies the impact of that delay. For example, if the leaking position is at position a, and a is also the position explored, then the cost of digging a is $50k, and the corresponding payoff is -50k. If a is the leaking position, but position b is going to be explored, then the digging cost at b is $60k, the cost of delay is $300k, and the payoff is -360k.

Using our system, the problem can be modeled in the two-level structure: At the credal level, an evidential system is created consisting of 21 test-variables Testi (i=1, ..., 21) with frames {+, -}, 11 symptom-variables Sympi (j=1, ..., 11) with frames {yes, no}, one diagnosis-variable diagnosis with frame {a, b, c, d, e, f, g, ω}, and the relations among them; at the pignistic level, the structure is very simple for this problem: it consists of only one decision variable treat(ment) with frame {clean-a, clean-b, clean-c, clean-d, clean-e, clean-f, clean-g, noclean}, one random variable diagnosis and one utility valuation utility bearing on the two variables. The diagnosis variables at both levels are identical, in which the belief function can be transferred from the credal level to the pignistic one as probability. The graphical representation is illustrated in Figure 3.

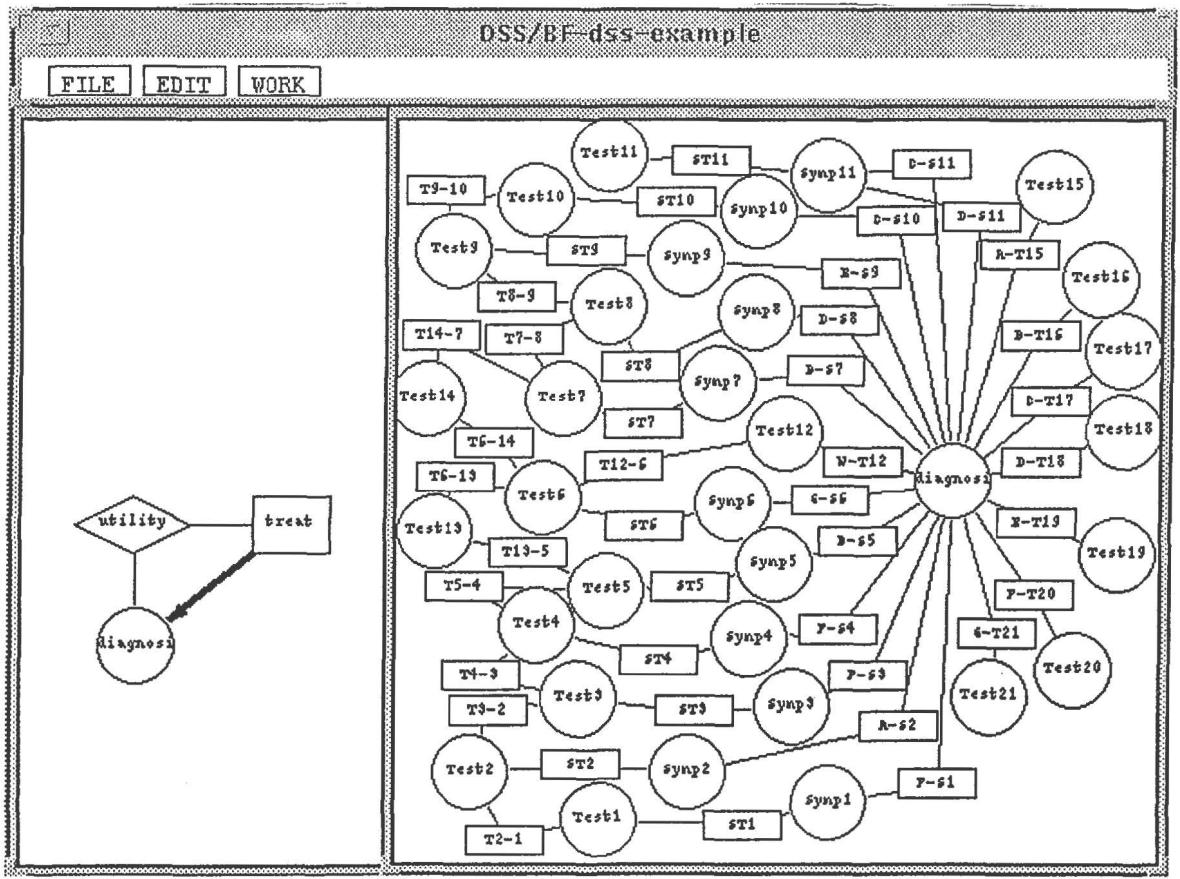

Figure 3: Graphical representation of the model

The above construction can be regarded as the structure-knowledge creation. To enter the beliefs in the system is a quantitative knowledge construction. From the above description of the problem, we can find that it would be easier for the users if the relations could be input under a conditional form (Smets 1991) instead of by entering joint belief functions. The system provides such a facility. In the example, for constraint (1), we can define a belief function for the relation variable STi (Figure 3) connecting variable Testi and Sympi (i= 1, ..., 11) through a conditional belief function input facility, as shown in Figure 4. Since in TRESBEL, belief functions to be propagated should be joint belief functions, the system will transfer each conditional form to the joint belief function, which is called as the ballooning extension of such conditional belief functions (Smets 1991). For example the ballooning extension of the conditional belief function in Figure 4 is joint belief function over the frame {yes no}×{+ -} (of $ST_i$) with mass 0.99 to the subset {(yes, +), (no, +), (no, -)} and 0.01 to the whole frame of $ST_i$; Similarly, we can define the belief functions in a conditional form for the other constraints.

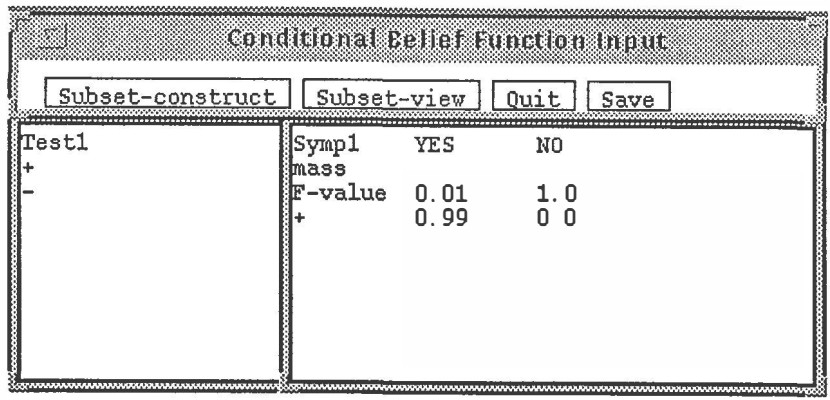

Figure 4: a conditional belief function input facility

After users commanding "work", the system will compute the optimal solution and prompt a window for showing the tree of suggested tests (Figure 5). As for the example, the system first suggests doing a test at point 12. If the test result is negative (-), then the next test is suggested being done at point 17; if the result is positive (+), then no more tests are needed, and so on. Furthermore, users could invoke any node of the tree for inquiring further questions. For example, At the node "Test 12", users can ask why the test is done at this point, the system will answer this question by showing and comparing the expected utilities of all the tests. The test is selected at point 12 is because the expected value of testing at 12 is the largest. Users can also ask the question about the consequent action - where to dig - according to the test. For example, After testing at point 12, and if the result is positive, then no test is suggested. At this moment, users can invoke the node "Notest" to ask what to do (where to dig) next. The system will suggest not digging at any point from **a** to **g** (i.e., "noclean" is optimal for the decision variable treat) since the leaking is coming from somewhere above this area. The level of the suggestion tree can be defined by the users, or by default the computation will be complete when all the leave node of the tree are "Notest".

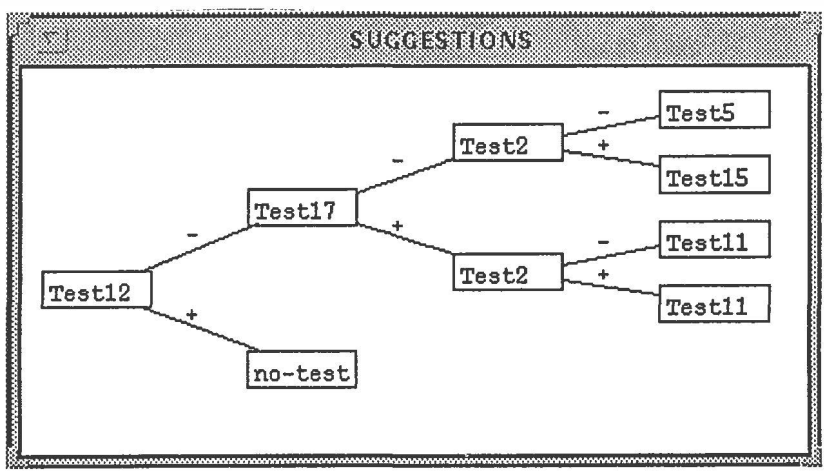

Figure 5: The tree of suggested tests for the example

## 3. CONCLUSIONS

The shell of a belief function-based decision support system has been described. The system is an integration of a VBS for Bayesian decision analysis and an evidential system for belief functions propagation, both on the context of the valuation-based framework. It is based on the theory of transferable belief model (Smets 1988). The system uses two tools: TRESBEL and VBSD (a successor of Pulcinella), and it uses the pignistic transformation to connect the two. It is designed and implemented in an interactive graphical way. A conditional belief function input facility for belief function input makes it easier for the users to construct the knowledge into the system. Furthermore, it provides a functional interface, making itself more flexible and easy-to-use. Further work on this system is continuing. Currently, the presented procedure carries out the optimization using a classical stepwise procedure. At a given moment, the best test according to the available information is selected, and the next best test is selected based on the result of the preceding test. The global optimum could, in principle, be found, but the combinatorial explosion problem makes it infeasible when the number of tests and the sizes of their frames are large. The presented procedure is in fact a general-purpose heuristic for the problem-solving. The future work will try to use other techniques of heuristic algorithms to select a better combination of tests efficiently.

### Acknowledgments

The authors would like to thank Prakash P. Shenoy for commenting on the preliminary version of this paper. This work has been developed within the MUNVAR project funded by the CEE-DGXII. It has also been partially supported by the DRUMS project funded by the Commission of the European Communities under the ESPRIT II, Basic research Project Action 3085. The first author would like to acknowledge the support in part from a grant of IRIDIA, Université libre de Bruxelles and in part from the General Research Fund No. 3605-XX-0038 of the University of Kansas. The authors also wish to thank the anonymous referees for their comments and discussion.